# Simulated Reasoning is Reasoning!

Philosophical Notions on a Technological Breakthrough

Hendrik Kempt[1], Alon Lavie[2]

[1] Chair of Applied Ethics, Department for Society, Technology, and Human Factors, RWTH Aachen University

[2] Language Technologies Institute, School of Computer Science, Carnegie Mellon University

Correspondence: hendrik.kempt@humtec.rwth-aachen.de

**Abstract**

Reasoning has long been understood as a pathway between stages of understanding. Proper reasoning leads to understanding of a given subject. This reasoning was conceptualized as a process of *understanding* in a particular way, i.e., "symbolic reasoning". Foundational Models (FM) demonstrate that this is not a necessary condition for many reasoning tasks: they can "reason" by way of imitating the process of "thinking out loud", testing the produced pathways, and iterating on these pathways on their own. This leads to *some form of reasoning* that can solve problems on its own or with few-shot learning, but appears fundamentally different from human reasoning due to its lack of grounding and common sense, leading to brittleness of the reasoning process. These insights promise to substantially alter our assessment of reasoning and its necessary conditions, but also informs the approaches to safety and robust defences against this brittleness of FMs. This paper offers and discusses several philosophical interpretations of this phenomenon, argues that the previously apt metaphor of the "stochastic parrot" has lost its relevance and thus should be abandoned, and reflects on different normative elements in the safety and appropriateness-considerations emerging from these reasoning models and their growing capacity.

## 1. Introduction

When in the 1980s and 1990s personal computers became mainstream technological devices in most people's homes, their reception and interpretation was muddied with anthropomorphism and misunderstandings. The processing noises of the hard drives were described as "the computer is thinking", while people ascribed different moods to the device depending on daily performances (Reeves & Nass 1996), often without representing the implications of such statements. Since then, the computational capacities of personal computers have grown exponentially, and algorithms have emerged that are learning machines.

With the advent of widely accessible general-purpose chatbots in 2022, the reception and interpretation of these machines has been once again beset with anthropomorphic projections and interpretations that are divorced from the factual technological background. This led to emerging narratives of these machines and their prowess being tinted with an overly optimistic outlook: the imminent arrival of artificial general intelligence within a few years was promised, which were fueled by hype and the (intentional) over-selling by executives and tech-enthusiasts of the trajectory we are on.

As an important counterpoint and interpretative tool, Bender et al. (2021) suggested a metaphor characterizing the functions of LLMs and other foundational models more precisely: they proposed that these algorithms are more akin to "stochastic parrots" than human speakers.

In this paper, we critically examine both of these interpretations in the light of more recent state-of-the-art reasoning models. Reasoning models exhibit the ability to provide a reasoning explanation of their output and progress through this explanation in refining their output - i.e., based on their intermediate outputs, they can sequentially "reason" through the problem they are solving. Our guiding thesis is that these models are proving that there is a crucial difference between mere pretrained LLMs, that might be considered stochastic parrots, and these more advanced models, although they still fall short of any established well-understood interpretation that ascribes more substantial internal representation to their "reasoning". Thus, a new philosophically informed description of their processes is needed, which we call "simulated reasoning". This simulated reasoning should count as a subset of the complete reasoning processes humans perform. However, simulated reasoning is characterized by certain limits, which we aim to explore in this paper.

To substantiate this thesis, we proceed as follows. **First**, we introduce and characterize contemporary reasoning models and their features. **Second**, we discuss how the metaphor of stochastic parrots has been successful at characterizing previous models and why it might now face limitations. **Third**, we introduce and discuss several available interpretations of the process in question. This includes our own proposal - that simulated reasoning can be a form of reasoning - and describe some of its features and distinctions versus non-reasoning LLM inference processes and more substantial ("human") reasoning. **Fourth**, we suggest that reasoning machines ought to be treated differently moving forward with respect to their safety, appropriateness considerations, and other normative and regulatory requirements. This includes new opportunities for improved safety measures, as well as novel challenges to the safety-architecture put in place over the past years.

## 2. A short State of the Art of Reasoning Machines

Foundational Models (FM) have seen remarkable improvements over the last few years, with new benchmarks being met and surpassed on a regular basis. Despite uncertainty about whether future achievements will progress with the same pace due to doubts about the so-called "scaling laws" (Bahri et al. 2024), the current state of AI and its emerging capacities are reasons for philosophical reflection on ethical and epistemological levels. For a long time, the transformer-based architecture, the dominant engineering paradigm for AI, was considered limited in its ability to achieve human-like reasoning. A common belief, particularly among philosophers and some computer scientists, is that human-level reasoning requires an understanding of concepts and their relationships, i.e., a grasp of symbols. The manipulation of these symbols was seen as the only path to comprehension of the kind humans are capable of (Winston 1992, Franklin 1995).

However, the advent of what has been called "reasoning models" (with OpenAI's o1 counting as the first of its kind (Edwards 2024)) has shown these assumptions to be more fragile than previously thought. These models, developed by organizations like OpenAI, Google, Anthropic, and DeepSeek, demonstrate an ability to solve complex problems by seemingly "thinking through" them, a process that appears similar to human step-by-step reasoning. While FMs, like other Large Language Models (LLMs), do not "understand" or "know" in the philosophically demanding sense of possessing mental states or representations, they can produce a prose that explains their reasoning steps, much like a person would. They simulate a textual verbalization of the process of sequential human reasoning, and through this emulation, they manage to successfully navigate tasks that were once the exclusive domain of human intellect.

The primary innovation that separates these reasoning models from their predecessors lies in their sophisticated training processes. Earlier transformer-based LLMs were largely trained to predict the next word in a sequence, which, while effective for generating coherent text, did not explicitly teach them to reason. The new generation of models undergoes a multi-stage training regimen designed to instill this capability. Such training often begins with **Supervised Fine-Tuning (SFT)**, where the model is trained on a high-quality dataset of problems and their detailed, step-by-step solutions composed by humans (Ouyang et al. 2022). This teaches the model to imitate the format of a reasoning process. However, the jump in sophistication and problem-solving abilities comes from subsequent stages that use reinforcement learning to refine the model's abilities.

A key technique here is **Reinforcement Learning from Human Feedback (RLHF)**. In this phase, model-generated solutions are ranked by humans for their quality and logical coherence (Christiano et al. 2017). A separate "reward model" is then trained on these human preferences to act as a scalable, automated judge of what constitutes a "good" reasoning process. The LLM is then further fine-tuned using reinforcement learning algorithms, such as *Proximal Policy Optimization (PPO)* (Schulman et al. 2017), to maximize the score it receives from this reward model. This process incentivizes the model not just to produce a correct final answer, but to generate a logically sound and human-preferred path to that answer. More recent and simpler techniques like *Direct Preference Optimization (DPO)* (Rafailov et al. 2024) achieve similar results by directly optimizing the model on the preference data itself, bypassing the need for a separate reward model.

Further extending this paradigm, techniques like *Reinforcement Learning from Verifiable Rewards (RLVR)* (Lambert et al. 2024) have been introduced. With RLVR, the model's reasoning is not just judged by human preference, but by whether its intermediate steps or final answer can be programmatically verified for correctness. This is particularly powerful in domains like mathematics or coding, where an external tool or verifier can definitively check the validity of the model's work, providing a strong and objective training signal.

These training innovations have profound implications for the models' inference processes—that is, how they go about solving problems when prompted. They enable a technique often called *chain-of-thought (CoT) prompting* or "thinking out loud" (Wei et al. 2022), which leads back to the initial observation that reasoning models seem to imitate human ways of reasoning.

Instead of attempting to produce an answer in a single step, the model is encouraged to generate a series of intermediate reasoning steps. This allows the model to break down complex problems into more manageable sub-problems, and critically, it allows the model to use its own generated text as a scaffold for further reasoning. It can test pathways, identify its own errors, and iterate towards a solution, a process that appears fundamentally different from the simple text prediction of earlier LLMs and more akin to genuine problem-solving. Recent studies (Merrill and Sabharwal, 2024) have also established that the class of problems solvable by transformers complemented by inference-time "Chain-of-Thought reasoning" computation is in fact a proper superset of "single-step" transformers. These differences have vast implications on both what Chain-of-Thought reasoning models can (and cannot) do, as well as on how to monitor and control their behavior, particularly with respect to safety.

Thus, these models demonstrate that reasoning can be learned and performed as a behavior, by imitating and then refining the process of successful reasoners, without necessarily possessing a complete underlying "understanding" of the concepts involved. They show that the simulation of a human reasoning process can itself constitute a valid form of reasoning, capable of producing novel information and solving problems on its own. This shift from models that merely retrieve and recombine information to ones that actively reason through problems marks a significant milestone, compelling a re-evaluation of our understanding of reasoning itself.

## 3. Philosophical Thoughts on Stochastic Parrots

A popular metaphor to characterize the capabilities and limits of FMs, and especially LLMs, has been the one of "stochastic parrots" (Bender et al. 2021): due to their architecture, language models are ultimately stochastically processing the likelihood of tokens, usually words, in their text generation and thus output - that is, their output is a sequence of highly educated and probable guesses. Akin to a parrot, who repeats and potentially recombines sounds (i.e., words) they pick up from their surroundings, they can sound very much like human speakers without understanding the contents of their utterances, and thus not being bound by truth or facticity while speaking with veracity.

This mismatch between the process of text-generation and the contents of said text can explain the occurrence of "hallucinations" or "confabulations", in which the FM produces output that appears linguistically correct but is factually inaccurate. As they merely guess what an appropriate output would appear like, LLMs are routinely producing utterances with claims to veracity without recourse to explicit grounding with the real world. Since hallucinations imply mental states, and confabulations imply an assertiveness behind the utterances, both have been discussed critically (Hicks et al. 2024). The analytic category of "bullshitting" (Frankfurt 2005), i.e., uttering without regard to truth or falsity, has been applied to these errors as well, although not uncontested. The reason here is that even bullshitting, like almost any other speech act, contains some intent or strategy (with bullshitting, the intent is to advance and promote the output to a given conversational, .i.e., contextual, outcome). These conceptualizations of the lack of truth-boundedness of even very advanced LLMs still match the basic characteristic of stochastic parrots, however: Parrots do not utter for the contents of their utterances as they do not understand their semantic content, but merely for their signalling effect to others.

By now, FMs appear to be excellent imitators of human speech, with an architecture generally accepted to not be reasoning in the human-like sense. This has led philosophers and computer scientists to propose abandoning the Turing Test as a benchmark for intelligence, as stochastically guessing well is generally accepted not to capture the intelligence we mean when we think of human reasoning capacities. Johnson-Laird and Ragni (2023) have pointed towards the difference of reasoning that these machines perform in comparison to humans and suggest three different elements of tests that point towards differences between FMs and human reasoning. The first being psychological tests of reasoning, the second an investigation of the machine's own interpretation of its reasoning and third examining the source code (ibid., 2).

The thrust of the argument, however, lies in the fact that the metaphor of LLMs as stochastic parrots has reached its limits: it evokes the equation between two types of speech - that is "parrot speech" and LLM-speech. It suggests that LLMs utter sentences by chance, without an appreciation of the context in which these utterances fall and that it does not account for the semantic changes to the common ground laid out by those utterances.

Such an equation, however, is underestimating the capabilities of the latter: LLMs, as a matter of fact, are more sophisticated than highly skilled parrots. There are two reasons for being skeptical of the continued aptness of the metaphor to describe LLMs.

First, the capacity of LLMs has grown enough to outgrow the metaphor's point: it does aptly characterize the fundamental difference between human and LLM-speech, but at the expense of limiting LLM-speech: instead of creating a new category of assessing LLM-sophistication, it boxes both parrots and LLMs into a "fundamentally not human-like"-category. However, ontological determinations are not determinations about prowess: the fact that LLM-speech is fundamentally different from human speech does not require it to be fundamentally worse, which seems to have been the original point of the metaphor.

Second, in that vein, warnings about the limits of those models, despite their appearance as proper speakers, asserters, and reasoners, can have detrimental effects on the wider public. If someone with little knowledge about LLMs is guided by the metaphor of parrots, they might underestimate the capacities of LLMs and thus become more vulnerable towards the risks associated with their use. It also might dampen the demand for regulation or limitation of these models if their image is one of parrots.

Thus, reasoning models ought not to be conceptualized as mere stochastic parrots anymore. As outlined in section 1, they manage to learn from their own semantic output. Given the advancements made in the field, we should expect further improvements through combinations with knowledge-graphs or other additions to the validity of their reasoning or soundness of their conclusions. Such progress begs the question what to expect next.

The comparison should no longer be whether their reasoning is "human-like", but rather whether it is "human-level" or of some other kind of capacity or benchmark. FMs appear (soon to be) capable of reliably reasoning through complex problems that previously were limited to human-reasoning, and in such a robust way that they can be used in wider contexts. They manage, unlike parrots, to gain information from their own utterances, and yet do so differently than human reasoners.

These issues make a renewed account of what kind of speech these LLMs are producing urgently needed. In the following, we draw a sketch of how we should categorize this "simulated reasoning": that is, as a form of human-level reasoning.

## 4. Simulated Reasoning is Reasoning

So far, our analysis suggests that Bender et al. (2021) were correct in pointing towards the lack of understanding and reasoning-capacities of early language models. However, given the developments outlined in sections 1 and 2, the aptness of the metaphor of a parrot has reached not only its limits, but also might become detrimental in the education of the wider public to appreciate the capabilities of those machines.

We thus should offer a different philosophical understanding of what is being performed in these machines that accounts for these new developments and offers a path forward. For this, we should acknowledge first some basic observations: reasoning LLMs are apt at some types of reasoning that humans are also deploying on a regular basis: inductive reasoning and abductive reasoning appear both easy ways for LLMs to perform based on their training. As learning machines, training data provides instances of connections that create pathways that lead to inductive reasoning. Learnings of a machine are largely achieved by showing it what successful instances of certain events look like. Given a task, then, reasoning models can reason from those instances onto the next ones.

### 4.1 Deduction and Computation

Inductive reasoning, although easily and often deployed by human thinkers, does not constitute human reasoning in full. That is in spite of how much of our learning is being done by "learning by doing" or learning by examples - which both are modes of inductive reasoning. What is missing is that we are also deductive reasoners, i.e., we deduce consequences from principles, and measure our expectations against what those principles predict.

Felin and Holweg (2024) point out that a purely statistical approximation of reasoning is not enough to qualify as reasoning. Their explanation for the problematic assumptions in these contexts is the lack of distinction between "computation" and (causal) "reasoning". i.e., the difference between the relation between inputs and outputs (computation) and the way humans interpret the world (by theory-laden causal inferences). As causal reasoners, we have expectations of events occurring and processes unfolding in the world that are a priori. On this basis, we can have modal logic in which certain occurrences in the world are impossible, and thus assertions that make impossible claims must be false. LLMs lack "experience" in a more complete, multi-modal, multi-sensory sense (i.e., grounding), are usually thought to not have an internal "map of the world", and thus cannot create causal expectations and inferences about that world. Hence, no strict deduction can be achieved. Felin and Holweg call this the "data-belief asymmetry", as the former cannot achieve the same reasoning capacities as the latter. Thus, LLMs must by principle remain incomplete or wonky reasoners.

This also explains how some of the common-sense mistakes still occur: as causal reasoners, we can distinguish between cases that might appear structurally or semantically similar at first, but that cannot pass the test of causal reasoning. Without an innate mode of causal thinking, however, the similarity between cases can blur to the degree that a clear reasoning process goes awry. The mere structure of an input, thus, produces the structure of an output that is not causally sound.

Thus, they conclude that data-belief-asymmetry is the best explanation for the differences between AI-based reasoning and its human counterpart (ibid.).

Van Rooij et al. (2024) complement this finding in their analysis and demonstrate that the computational approach to reasoning will *formally necessarily* lead to impossible resource demands to advance further. They do not conclude that cognition cannot be fully explained in computational terms alone, and thus human cognition may theoretically be reducible to a computational theory. However, they prove that the remaking of such computational processes is "intractable" even under idealized conditions. This renders the recreation of human-level and human-like cognition mathematically impossible within a computational framework, i.e., in which contemporary AI is being developed.

Felin and Holweg (2024) remain somewhat unclear on how they ultimately consider AI cognition and reasoning to be related, and van Rooij et al. (2024) pursue a slightly different goal in their mathematical proof of refuting the practical computationalism for human-level/-like cognition. However, seeing that reasoning models are capable of all sorts of cognitive maneuvers, including self-correction (either by intermediate assessments of their outputs, or by being pointed towards the issue by users), demonstrates that their capacity might be more substantial than what the authors could imply, and that the method of creating reasoning machines is still worthy to be taken seriously. While many authors who reject the possibility of current AI-paradigms achieving human-like reasoning acknowledge that this is the case, merely pointing towards the substantial difference between human and AI-cognition may not satisfy and lead, similar to the stochastic parrot metaphor, to a structural underestimation of these machines' capacities.

4.2 Fuzzy Reasoning?

There are means to support the thesis that reasoning models are reasoners in a more complete sense than suggested by these and other authors. LLMs are usually understood as probabilistic reasoners, i.e., their output is a function of probability. Considering that their utterances are not connected to a causal reasoning but to probabilities, they remain incapable of logically valid and sound deductive reasoning.

Fuzzy logic might be a plausible way of approaching deductive reasoning strategies with data inputs alone. This has been demonstrated in clinical contexts, where LLMs were used as "fuzzy judges" (Zheng et al. 2025) to make judgments on fuzzy propositions. Treating truth values as ambiguous or vague rather than uncertain could explain the difference of reasoning strategies detectable here: The use of LLMs as "fuzzy judges" could give rise to an interpretation of reasoning models as "fuzzy reasoners", i.e., an interpretation that posits that their grounding is not based on probabilities, but on vagueness resulting from a lack of sensory perception and common-sense.

## 4.3 Hidden Neurosymbolism?

Felin & Holweg (2024) claim that the lack of causal reasoning and proper grounding is the explanation for a lack of comprehensive reasoning capacity in reasoning models. If, however, reasoning models over the coming years should reach a level of reliability that is akin to causal reasoners, their identified asymmetry between data and beliefs might not be sufficient an explanation.

The fact that the reasoning of reasoning-models can lead to results similar or often same to deductive reasoning affords an ontologically more weighty interpretation of this process: that the reasoning-reports of those models are actual verbalizations (or semantic mappings) of the computational process underneath, in which the same reasoning processes have been going on. In this sense, FMs appear closer to be doing symbol manipulation, as the processes, of which the reasoning chain is thought of as a mere verbalization, do attest to an understanding of the subject matter. Such an explanation could also have consequences for our own reasoning, as we might also merely verbalize what is going on on a neural level.

There are indications that reasoning models begin to form short cuts that do not represent natural language, but are assigned semantic meaning through heuristic repetition (Korbak et al. 2025): reasoning models appear to create their own language to find more efficient pathways of arriving at the desired result. Thus, there is at least some initial evidence that these models might be able to create quasi-computational cognitive shortcuts that are distinct from the training data and the reinforced learning. In this sense, verbalizing in human language what those shortcuts could stand for might be a viable interpretation.

However, for many instances of the reasoning that is already successfully performed such an interpretation appears to be computationally and ontologically implausible. This means that in order for this interpretation to be possible, we also have to assume that the LLM is in fact able to precisely map semantic and logical notions on their own computational processes, and that those described shortcuts are not mere correlations but closer to symbolic understandings. While this is not impossible, we currently have limited reason to believe that LLMs have advanced that far. However, it is not impossible. While symbolic AI, as previously understood, suggested an understanding of the symbols and thus enabled manipulation, the idea that algorithms are "learning machines" suggests that statistics may hold sufficient precision to perform these tasks, too. This could be found in *neurosymbolic* accounts of AI, in which the combination of deep learning and symbolic reasoning is combined (Sheth, Roy & Gaur 2023). Whether we have witnessed this here, however, is an open question.

### 4.4. Simulated Reasoning

Lastly, we offer a view on these reasoning machines that is not comparing human-like and machine-reasoning with the intent to show the difference. Rather, we take a phenomenological perspective and first describe what is happening: the reasoning models appear to do what a human process of "thinking through a problem" would typically look like. It does make mistakes, and is easily tripped by improper inputs or jailbreaks, making the entire endeavor brittle and not very reliable. However, it is effectively capable of learning from its own output, and thus self-correct and self-improve. This is very much unlike parrots have been using imitations of human speech to create statements.

Without any further information about the system, we could thus classify this reasoning as some kind of reasoning. Reasoning, understood as an ability to solve problems or produce new information by virtue of one's own intellectual capacity, can, in fact, be purely behavioral. This means that the simulation of the human reasoning process, which consists in imitating the step by step reasoning humans do to "think through" a problem, can constitute reasoning in its own right, i.e., it can produce new information on its own. While we have been taught in our reasoning to evaluate our steps, often with conversational scrutiny from others, reasoning models usually present their full elaboration of their reasoning without the ability for users to correct in the process. This renders some of the reasoning results off, as one wrong step in the simulated process will trickle down to the output. However, this demonstrates that reasoning can be learned by imitating other (successful) reasoner's behaviors, without necessarily having to understand the processes behind the behavior (cf. Venhoff et al. 2025), in part even with results outmatching median human reasoning abilities (Moore et al. 2025).

While this is reminiscent of the data-belief asymmetry, it draws the rather opposite conclusion: some humans might be excellent causal reasoners that have reliable means for thinking through a practical or strategic problem that will not be replicated with simulated processes. However, most people are most of the time for most purposes not skilled like Aristotle or Newton or Terence Tao were and are in their fields or reasoning. As a matter of fact, some parts of our thinking processes, called "fast thinking" by Kahnemann (2011), are similar to this kind of simulated reasoning exhibited by reasoning models.

In a fair comparison to reasoning models, many people's predictions and expectations about the world they care about are formed by experience in a general frame of what is possible. It appears odd to ask reasoning models, still in their infancy, to tackle problems most people would not be able to reason-through without flaws either.

Take another dimension of this argument: practical dimensions of understanding do often imply a "knowing how", i.e., the knowledge of how to do something, rather than mere "knowing that". This can be pure muscle memory, heuristic processes, or other non-cognitive ways of mastering a task. Here, most reasoners faced with the question of how to solve a practical problem rely on their own acquired experience that might not be propositionally represented, but merely assumed based on previous experience. In this vein, some authors argue that reasoning models exhibit "abilities" that come with having a theory of mind (e.g., Moore et al. 2025). While we agree with Felin & Holweg (2024) and van Rooij et al. (2024) that this is not sufficient to conclude a theory of mind to be present, our understanding of "behavioral reasoning" being a valid subset of reasoning-capacities can explain these reasoning-abilities and take them philosophically seriously.

## 5. Different Reasoning, Different Safety

Conceptualizing machine reasoning as different from some, but similar to other forms of human reasoning, leads to different measures to be taken to ensure the safe development and deployment of reasoning machines moving forward. These different measures are representing opportunities to make FMs more robust, more safe, and less prone to adversarial attacks or inappropriate utterances. At the same time, their way of operating also presents genuinely new challenges or amplifies existing limitations with LLMs that also should be reflected on within a normative analysis. As with previous iterations and generations of LLMs, this field ought to not be left to that label of "AI safety" alone, as safe AI is not being normatively sufficient (cf.

Kempt, Lavie & Nagel 2024 for an analysis for the lacking "safety normativity").

In the following we will discuss 1) the opportunities for improved safety by way of sequential computation, 2) the boundaries set for these models, and how their reasoning capabilities may be used to strengthen or undermine those boundaries, 3) the robustness of reasoning models and how the continued lack of common sense may affect safety precautions, 4) execution plans and the safety-concerns emerging from reasoning models that lack grounding and yet can formulate real-world plans.

5.1 Improved Safety and Opportunities

The consequence of reasoning as a sequential process of inferences provides several opportunities to improve the safety and appropriateness of the output produced. This is largely for three reasons: first, safety and appropriateness considerations are not only limited to be implemented in the training phase, as is required by non-reasoning models. While non-reasoning LLMs have some standard safety-procedure fine-tuned into their process, the product of that process, then, is not correctable even if it violates the intended safety measures.

At best, a filtering system in the output-phase can stop the output and replace it with a safety-notice, but it cannot correct or adjust its output according to safety- and appropriateness considerations. With sequential computation during inference, reasoning-models now can dynamically control themselves in that process and thus are more adaptive to those normative concerns. The ability to adjust outputs within the sequences of inference allows for more specific safety- and appropriateness measures to be installed, as a reasoning model can test for unintended or unforeseen output.

Second, the sequential nature of reasoning - as outlined in section 3.4 to be behaving like human reasoning without being so - can be understood as a way of self-correction that may lead to continuous self-improvement of the output of reasoning models, and thus to improved safety. In this sense, it can also be considered a form of "meta-reasoning", as it not only supervises its own output, but rather can test/compare that output to verified knowledge via RAG.

Third, it is now possible to apply another separate model specifically designed to monitor and check the accuracy and/or consequences of a reasoning process of a given model - also at inference time. While these methods are admittedly potentially costly and not fail-safe (Zhao et al. 2025), they provide several opportunities for safety checks during the inference phase that can correct or adjust the output.

The possibilities to combine either several reasoning models, or to train one model to constantly self-improve, suggest that reasoning models not only bring new safety concerns, but actually also new safety opportunities to ensure generative AI to be more safe to use, more appropriate in its outputs, and more trustworthy in its implementation.

### 5.2 Setting Boundaries and Jailbreaks

Increased reasoning-capacity implies the ability to also better reason about problems that might be impermissible to solve. The idea of improved safety and appropriateness opportunities from the previous section is mirrored by the fact these are necessary to limit reasoning models. The better the reasoning capacities, the more dangerous it can become to reason about certain topics. While in controlled and open contexts, such as research, this capacity holds immense promise to improve key aspects of enhancing researchers' abilities to form innovative reasoning pathways, develop novel ideas and methods, and improve our means of discovery and knowledge-management; and yet, the economic truth of the matter is that many reasoning-models are so costly that only a widespread adoption may return the investments made. It is likely that reasoning-capabilities will be used in problematic contexts, with undesirable purposes.

The concerns (and subsequent necessity for apt safety measures) are aimed at both the question of **method** as well as **subject**: a reasoning AI may be used to reason about its own or other AI's boundaries, their methods, and ability to jailbreak them, allowing users a relatively simple path towards unsafe, inappropriate or otherwise impermissible output (cf. Hagendorff, Derner & Oliver 2025). This is a distinctly new challenge to previous "conventional" LLMs, as the reasoning may provide the key to some of these jailbreaks. Non-reasoning LLMs are still lacking proper safety-protocols that keep them "jailed", enabling them to perform highly unethical behaviors by creating harmful outputs. Jailbreaking reasoning models could lead to even worse results.

Worse yet, it can also also be used to reason through complex matters that might be abused for malicious purposes. In this sense, reasoning models become subject of dual-use concerns, as the application of its capacity for malicious purposes is difficult to restrict by design.

Lastly, with reasoning models we have observed their tendency to create reasoning-pathways that are not represented in natural or human-made coding languages, but that are shortcuts of the model's own making (Pfau et al. 2024). These shortcuts are unintelligible for humans. This creates what is called the "control problem" of AI, i.e., the inability for human control over those processes due to the lack of interpretability of the decision pathways. This is not to say that we face the control-problems projected by AGI-theorists like Bostrom (Bostrom 2014) or Kurzweil yet. However, on a smaller scale, similar issues arise that are worth considering here: without the ability to appreciate the reasoning pathways of an AI, we might want to remain cautious in their implementation.

The proposed "chain of thought monitorability" (Korbak et a. 2025) represents an attempt at regaining traction of the model's navigation, though one should ask whether "monitoring" processes presents a sufficiency-criterion for acceptable risk- and safety-assessments. A similar issue arises in chain-of-thought reasoning where some meaningless filler-tokens can replace the verbalized reasoning process, achieving the same result (Pfau et al., 2024). However, these filler tokens appear to be necessary, as the task could not have been performed without either a filler token or the chain-of-thought reasoning. This also adds to the likelihood of non-human computation

5.3 Robustness vs. Brittleness

Next to previously discussed concerns of jailbreaks and lack of control, safeguarding the quality of outputs of LLM remains a challenge that only grows in relevance. Human reasoning is beset with biases that can be traced to our evolutionary requirements and the general human condition. This has made the reasoning humans employ in many ways robust but open to biases and inherent limitations. As we have elaborated in section 3.4, Daniel Kahnemann's work on different forms of thinking illustrates this (Kahnemann 2011): "fast thinking" rests on a robust set of experiences and intuitions, allowing us to make quick, commonsensical judgments. However, it is prone to certain types of bias that limit their generalizability, their applicability, and makes us vulnerable to exploitation or error. Slow thinking, on the other hand, is characterized by its methodical, reflective nature: in capturing general principles from which abstracted assumptions can be made, we can reason on solid logical, causal grounds rather than on intuitive, heuristical ones. The ability to "slow reason", thus, provides us with the intelligence to create, maintain, navigate, and also reject highly complex systems.

It is an open question of whether reasoning models will employ a similar bias of "fast thinking". They are not expert-systems executing a pre-determined decision-tree but rather, still similar to stochastic parrots, work with probabilities. There are methods of increasing the factuality of reasoning models, like retrieval augmented generation (RAG) which accesses knowledge-bases to ensure the output is "fact-checked" (Lewis et al. 2021). This may keep some of these biases at bay. However, the reasoning on more contextual, innately human topics, such as behavioral psychology, or those requiring practical knowledge not conveyed by textual sources may lead to biased results that appear as a result of logical reasoning, even though those results might be incredibly off the mark. This kind of brittleness of intelligence (Floridi 2023) will remain a challenge to these kinds of models, absent common sense and the ability to check answers not only based on retrievable knowledge but on plausibility within the wider context of a speech situation.

The lack of common sense and "understanding" of the reasoning the LLMs can produce, thus, present different issues than previously discussed and require a substantial revisitation of the AI ethics and safety literature.

5.4 Execution Plans

Next to the challenges of ensuring safety and quality of output, we ought to contend with the fact that reasoning models may form not merely linguistic plans, but plans that demand an execution in real-world contexts. Such plans thus expand the reach of LLMs from the two kinds of limitations we currently have in place (through the limitation to linguistic-representative interactions as well as those taking place exclusively within a textbox) towards real-life systems. Work in this area aims at creating LLMs that are able to make long-term plans that are sensitive towards environmental challenges of their realization (Erdogan et al. 2025).

These execution plans become especially salient in the challenge of the aforementioned decreasing monitorability: if the already beginning process of reasoning models finding their own pathways that do not work with intelligible human language deepens, we might not be able to predict or reign in models in their creation of execution plans. Out-of-control execution plans must not, however, be accompanied with malintent or a hidden agenda of an AI, but translates to direct practical issues: an out-of-control AI, solely focused on fulfilling the task given to it, could aim at realizing this task at any other cost (which is in Bostrom's famous example an AI tasked with creating paperclips, and using all available resources to that end (Bostrom 2014, p. 128-129)).

While we cannot reasonably draw strong normative conclusions about the future of those models, we should be aware that reasoning models have made the control problem a more salient issue than what it was previously.

## 6. Conclusion

We suggest that some parts of our understanding of the phenomenon of "reasoning" should be reconceptualized. This is to account for the fact that some forms of reasoning can purely be based on the imitation of reasoning-principles. This might not be a new insight coming from other disciplines - such as cognitive science and development: The growth of reasoning capabilities from infants to adolescents and to adults demonstrates that many of our ways to reason are not born from first principles (i.e., "top down" from a theoretical understanding of one thing towards another), but from a "bottom up"-way of learning by doing (i.e., learning by imitating other's behaviors). In that sense, some reasoning is best described as a behavior for which intentionality, understanding, and other elements are optional (although certainly useful).

This also tracks with different levels of expertise, education, and routine in reasoning processes: we sometimes do use heuristics and shortcuts to reason through a problem without fully grasping the underlying principles - we merely learned how to use them. It appears that LLMs have managed to do that, too.

The philosopher Annette Zimmermann once contended that "if you can do things with words, you can do things with algorithms" (Zimmermann 2021). This is holding true ever more: we reason with words, and even if one does not understand the words and proposition they form in the sense of a symbolic representation, we apparently can create machines that reason with these words as well. These ways of reasoning, albeit appearing similar to ours, occasionally even eerily so, are fundamentally different: without understanding, some reasoning processes can only be done with great effort or not at all, and still remain brittle and subject to common sense errors. And yet, many reasoning tasks can be performed without understanding, rendering the popular metaphor of stochastic parrots moot: they can, for many intents and purposes, speak, as they can correct themselves based on their own reasoning.

This has opened the door for new forms of AI-safety opportunities and concerns that ought to be philosophically, legally, and technologically reflected and further researched. The loss of monitorability of chain of thought reasoning could be a vital safety issue moving forward in LLM-capacities and their supervision and regulation.